\documentclass{article}

\usepackage[preprint]{neurips_2025}

\usepackage[utf8]{inputenc} 
\usepackage[T1]{fontenc}    
\usepackage{hyperref}       
\usepackage{url}            
\usepackage{booktabs}       
\usepackage{amsfonts}       
\usepackage{nicefrac}       
\usepackage{microtype}      
\usepackage{xcolor}         

\usepackage{amsthm,amsmath,amssymb}
\usepackage{mathrsfs}

\usepackage{algorithm}
\usepackage{algorithmic}
\usepackage[algo2e]{algorithm2e} 
\usepackage{amsmath}

\usepackage{graphicx}

\title{Zero-Shot Video Restoration and Enhancement with Text-to-Image Latent Diffusion Models and Multi-Modal References}

\author{
    Cong Cao\textsuperscript{\rm 1},
    Huanjing Yue\textsuperscript{\rm 1},
    Xin Liu\textsuperscript{\rm 2},
    Jingyu Yang\textsuperscript{\rm 1} \\
\textsuperscript{\rm 1}School of Electrical and Information Engineering, Tianjin University, Tianjin, China\\
\textsuperscript{\rm 2}Computer Vision and Pattern Recognition Laboratory, School of Engineering Science, \\ Lappeenranta-Lahti University of Technology LUT, Lappeenranta, Finland\\
\texttt{\scriptsize{\{caocong\_123@tju.edu.cn, huanjing.yue@tju.edu.cn, linuxsino@gmail.com, yjy@tju.edu.cn\}}}  \\
}

\begin{document}

\maketitle

\begin{abstract}
Zero-shot image restoration methods with text-to-image latent diffusion models have achieved great success in universal image restoration tasks without training. However, applying them to video restoration will result in severe temporal flickering. In this paper, we propose a novel framework for zero-shot video restoration and enhancement which uses a text-to-image latent diffusion model and multi-modal references. Through the proposed dual prompt tuning inversion and sampling, the inference time can be reduced to nearly 1/3 of the original. The performance and temporal consistency can be also significantly stregthened. By using the proposed texture-aware video token merging, the temporal correlation between frames can be further utilized to improve the temporal consistency. We futher propose the referenced self-attention and referenced token merging to support image reference. Experimental results demonstrate the superiority of the proposed method in restoring and enhancing temporally consistent videos.
\end{abstract}

\section{Introduction}
\label{sec:intro}

Recently, Denoising Diffusion Probabilistic Models (DDPMs) \cite{dhariwal2021diffusion} have shown advanced generative capabilities beyond GANs, which inspires people to explore diffusion-based restoration methods. Different from using supervised learning and a diffusion framework to train a model for a specific restoration task \cite{saharia2022image}, the works in \cite{song2019generative,lugmayr2022repaint,choi2021ilvr,kawar2022denoising,wang2022zero,chung2022diffusion,fei2023generative,rout2024solving,chung2023prompt,he2023iterative,rout2023beyond} utilize pretrained image diffusion models for universal zero-shot image restoration. Among them, \cite{rout2024solving,chung2023prompt,he2023iterative,rout2023beyond} work on latent space with text-to-image latent diffusion models.
However, there is no temporal modeling in pretrained text-to-image latent diffusion models to process videos. Although these methods have achieved promising results on image restoration, directly applying them to video restoration will result in severe temporal flickering.


Along with the appearance of powerful pre-trained text-to-image diffusion models like Stable Diffusion \cite{rombach2022high}, how to use text-to-image diffusion model for zero-shot video restoration/enhancement has garnered increasing attention. Generating temporally consistent results remains a key challenge in zero-shot video restoration/enhancement. In this work, we propose texture-aware video token merging to better balance temporal consistency and texture preservation. Another problem is accelerating the sampling process. For zero-shot video editing, DDIM inversion and DDIM sampling are common settings for acceleration. But they are not suitable for video restoration since the image for inversion is a degraded frame rather than a clean one. The combination of DDIM inversion and DDIM sampling biases the result towards the reconstructed degraded frame. Therefore we propose dual prompt tuning inversion and sampling to solve this problem. We optimize the conditional and unconditional embeddings in the inversion to better encode the information of degraded image and give a better initialization for sampling. Then we optimize the conditional and unconditional embeddings in the sampling to explore the potential of text-to-image latent diffusion model for video restoration. And our inversion and sampling can further improve temporal consistency by constraining the consistency of conditional and unconditional embeddings across frames.

On the other hand, existing zero-shot video restoration/enhancement methods did not consider introducing extra references \cite{zhang2019deep,lee2022reference} to guide the restoration/enhancement. Extra references, such as text (describing the ground truth video or preferred style) or other clean images (captured from different angles or obtained by retrieval), can provide additional information to assist with video restoration and enhancement. Our inversion and sampling strategy has already supported the text reference. To better support image reference, we further propose referenced self-attention and referenced token merging. By utilizing referenced token merging, the textures of the image reference can be transferred to the input degraded videos in a temporally consistent manner.

Based on the above observations, we propose a novel framework for Zero-shot Video Restoration and enhancement using Text-to-Image latent diffusion model and Multi-modal references (ZVRM).

Our contributions are summarized as follows
\begin{itemize}
\item[$\bullet$]{First, we propose a novel framework for zero-shot video restoration and enhancement using pretrained T2I latent diffusion model and supporting different multi-modal references, including no-reference, text-reference, image-reference, as well as both text and image references.}

\item[$\bullet$]{Second, we propose dual prompt tuning inversion and sampling for acceleration and text reference, propose referenced self-attention and referenced token merging for image reference, and propose texture-aware video token merging to preserve temporal consistency in video restoration and enhancement.}

\item[$\bullet$]{Extensive experiments demonstrate the effectiveness of our method in achieving temporal-consistent zero-shot video restoration and enhancement.}
\end{itemize}

\section{Related Work}

\subsection{Zero-Shot IR with Latent Diffusion Models}

Zero-shot image restoration methods are training-free, utilizing a pre-trained diffusion model as the generative prior and constraining the content of the result in the sampling process to be consistent with degraded images. According to the type of pre-trained image diffusion model used, zero-shot image restoration (IR) can be divided into pixel-space zero-shot IR and latent-space zero-shot IR. Pixel-space zero-shot IR \cite{song2019generative,lugmayr2022repaint,choi2021ilvr,kawar2022denoising,wang2022zero,chung2022diffusion,fei2023generative} utilizes pretrained unconditional image diffusion models working in pixel-space \cite{dhariwal2021diffusion}. Latent-space zero-shot IR \cite{rout2024solving,chung2023prompt,he2023iterative,rout2023beyond} utilizes pretrained text-to-image latent diffusion models, such as Stable Diffusion \cite{rombach2022high}. PSLD \cite{rout2024solving} is the first framework to solve zero-shot image restoration using a text-to-image latent diffusion model. \cite{chung2023prompt} proposes a prompt tuning method to jointly optimize the text embedding during sampling. \cite{he2023iterative} uses historical gradient information to guide the sampling process. However, all these methods are designed for image recovery tasks, and severe temporal flickering occurs when they are applied to degraded videos.

\subsection{Zero-Shot Video Editing}

With the recent success of text-to-image diffusion models like Stable Diffusion \cite{rombach2022high}, many works apply a pre-trained text-to-image diffusion model for text-driven video editing. The key is how to solve the temporal consistency problem. Rerender-A-Video \cite{yang2023rerender} proposes hierarchical cross-frame constraints to enforce temporal coherence in shapes, textures, and colors. TokenFlow \cite{geyer2023tokenflow} and VidToMe \cite{li2024vidtome} enforce temporal consistency by unifying self-attention tokens across frames. Inspired by these works, we propose texture-aware video token merging, which better balances temporal consistency and texture preservation in video restoration and enhancement.

\subsection{Zero-Shot Video Restoration and Enhancement}

ZVRD \cite{cao2024zero} proposes to utilize the pre-trained image diffusion model for zero-shot video restoration, and introduces the short-long-range temporal attention layer, temporal consistency guidance, spatial-temporal noise sharing, and an early stopping sampling strategy. DiffIR2VR \cite{yeh2024diffir2vr} proposes flow-guided video token merging to improve temporal consistency when applying diffusion-based image restoration models for video restoration. SVI \cite{kwon2024solving} converts the time dimension to the batch dimension and introduces a batch-consistent sampling strategy that synchronizes the stochastic noise components. VISION-XL \cite{kwon2024vision} further introduces a pseudo-batch consistent sampling strategy and pseudo-batch inversion to improve the performance. Although there are zero-shot video restoration and enhancement methods available, none of them consider the multi-modal reference \cite{zhang2019deep,lee2022reference} for video restoration. Multi-modal references, such as text (describing the ground truth video or preferred style) or other clean images (captured from different angles or achieved from retrieve), can provide additional information to assist with video restoration and enhancement. Our method is the first zero-shot video restoration and enhancement method that supports different multi-modal references, including no-reference, text-reference, image-reference, as well as both text and image references.

\section{Background}

\subsection{Latent Diffusion Models}

Latent diffusion models (LDM) operate in the latent space with an autoencoder $\mathcal{E}$, $\mathcal{D}$. First, an encoder $\mathcal{E}$ compresses the input RGB image $x$ to a low-resolution latent $z=\mathcal{E}(x)$. Then, the forward and reverse diffusion processes work on the latent, where the latent can be reconstructed back into an image $\mathcal{D}(z) \approx x$ by the decoder $\mathcal{D}$. In the forward diffusion process, Gaussian noise is gradually added to $z_0$ to obtain $z_t$ through Markov transitions with the transition probability.
\begin{equation}
\centering
q(z_t | z_{t-1}) = \mathcal{N}(z_t; \sqrt{1-\beta_t}z_{t-1}, \beta_t \text{I}),
\label{eq:ddpm_forward}
\end{equation}
where $\beta_t$ is the variance schedule for the timestep $t$. The backward process uses a trained U-Net $\varepsilon_{\theta}$ for denoising:
\begin{equation}
\centering
p_{\theta}(z_{t-1}|z_t) = \mathcal{N}(z_{t-1}; \mu_{\theta}(z_t, \tau, t),  \Sigma_{\theta}(z_t, \tau, t) ),
\label{eq:ddpm_backward}
\end{equation}
where $\tau$ denotes the textual prompt. $\mu_{\theta}$ and $\Sigma_{\theta}$ are computed by $\varepsilon_{\theta}$.

\subsection{DDIM Sampling}

Deterministic DDIM sampling \cite{song2020denoising} is employed to reverse diffusion process, which converts noisy latent $z_T$ to a clean latent $z_0$ in a sequence of timestep:
\begin{equation}
\label{eq: denoise}
    z_{t-1} = \sqrt{\alpha_{t-1}} \; \frac{z_t - \sqrt{1-\alpha_t}{\varepsilon_\theta}}{\sqrt{\alpha_t}}+ \sqrt{1-\alpha_{t-1}-\sigma_{t}^2}{\varepsilon_\theta} + \sigma_{t}\epsilon_{t}
\end{equation}
where $\alpha_{t}$ and $\sigma_{t}$ are parameters for noise scheduling \cite{song2020denoising}, $\epsilon_{t} \sim \mathcal{N}(0, 1)$. In practice, firstly $\boldsymbol{\hat{z}}_0$ is predicted from $\boldsymbol{z}_t$
\begin{equation}
\boldsymbol{\hat{z}}_{0} =  \frac{\boldsymbol{z}_{t}}{\sqrt{\bar{\alpha}_{t}}}-\frac{\sqrt{1-\bar{\alpha}_{t}} \varepsilon_{\theta}}{\sqrt{\bar{\alpha}_{t}}}
\label{x0}
\end{equation}
where $\bar{\alpha}_t=\prod_{i=1}^t \alpha_i$. Then $\boldsymbol{z}_{t-1}$ is sampled using both $\boldsymbol{\hat{z}}_0$ and $\boldsymbol{z}_t$
\begin{equation}
    z_{t-1} = \frac{\sqrt{\alpha_t}(1-\bar{\alpha}_{t-1})}{1 - \bar{\alpha}_t}z_t + \frac{\sqrt{\bar{\alpha}_{t-1}}\beta_t}{1 -\bar{\alpha}_t}\hat{z}_0 +  \sigma_{t}\epsilon_{t}
\end{equation}
where $\beta_t = 1-\alpha_t$. 

\section{Method}

\subsection{Overall Framework}

\begin{figure*}
    \centering
    \includegraphics[width=0.95\linewidth]{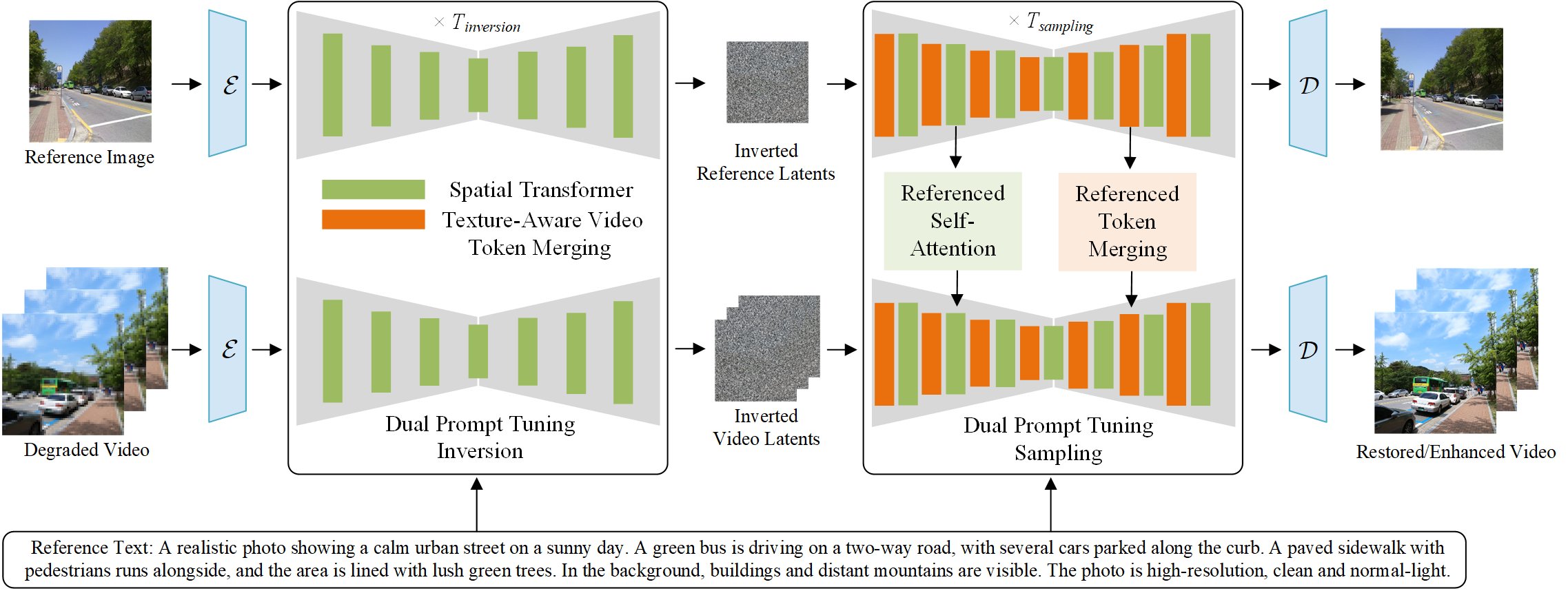}
    \caption{Overview of our framework.}
    \label{fig:framework}
\end{figure*}

Given a degraded video with $N$ frames $\{I_i\}_{i=0}^{N-1}$, our goal is to restore/enhance it to a clean video $\{I'_i\}_{i=0}^{N-1}$. Our method leverages pretrained text-to-image latent diffusion models (LDM) for video restoration. In LDM, a U-Net removes the noise of latent in the reverse process, which is constructed from layers of 2D convolutional residual blocks and spatial self-attention blocks. We replace all $3\times3$ 2D convolutions with inflated $1\times3\times3$ 3D convolutions, so that the network can process video. For better temporal consistency, we propose texture-aware video token merging. For text-referenced video restoration/enhancement, we propose dual prompt tuning inversion and sampling. Furthermore, we propose referenced self-attention and referenced token merging for image-referenced video restoration/enhancement. The framework is illustrated in Fig. \ref{fig:framework}. 

\subsection{Texture-Aware Video Token Merging}

Token merging has been proposed to speed up diffusion models \cite{bolya2023token} and improve the temporal consistency of generated videos \cite{li2024vidtome} by merging similar tokens within a frame or between different frames. However, these methods use a unified merge ratio for all pixels. For image restoration, different areas have varying levels of difficulty in restoration \cite{kong2021classsr,jeong2024accelerating}. Smooth areas are easier to restore than areas with rich texture. We find that in video restoration, areas with rich texture suffer more visual quality damage compared to smooth areas when using a larger merge ratio. Therefore, we propose assigning different merge ratios to smooth and texture-rich areas.

We utilize the classifier in \cite{jeong2024accelerating} to classify pixels of the input frame as belonging to either smooth or texture-rich areas. Since the input frame suffers from degradation, which affects the classification of different areas, we apply the classifier to the generated frame $\mathcal{D}(\hat{z}_0)$ in the second half of the sampling process every $N_{c}$ steps ($N_{c}$=25) to update the classification of different areas. For low-light enhancement, we brighten the input frame using a gamma correction process for classification.

\subsubsection{Texture-Aware Spatial Token Merging}

We split the overall token chunk $\textbf{T}$ of a frame into token sets $\textbf{T}_{s}$ and $\textbf{T}_{t}$, based on the smooth and texture-rich areas, respectively. For each token set, we further partition the tokens into a source set $\textbf{T}_{s}^{src}$ ($\textbf{T}_{t}^{src}$) and destination set $\textbf{T}_{s}^{dst}$ ($\textbf{T}_{t}^{dst}$), and compute the pair-wise cosine similarity between the source and destination sets. The $r_{s}$ ($r_{t}$) most similar paired source-destination tokens from $\textbf{T}_{s}$ ($\textbf{T}_{t}$) are merged. The merging ratio $r_{s}$ is much larger than $r_{t}$. After self-attention, the merged tokens are unmerged to maintain the original shape. The spatial merging and unmerging operations are the same as in \cite{bolya2023token}. The unmerged $\textbf{T}_{s}$ and $\textbf{T}_{t}$ are finally assigned to their original positions.

\subsubsection{Texture-Aware Temporal Token Merging}

For temporal token merging, we select the first frame as the key frame and merge the other frames into the target frame based on token similarity between frames. Texture-rich areas often become very blurry after merging with a high temporal token merging ratio, while smooth areas remain almost unchanged. Therefore, we assign a higher merging ratio to the smooth areas of frames than to the texture-rich areas to further reduce computational costs. Specifically, we merge $\textbf{T}_{s}$ and $\textbf{T}_{t}$ of different frames, respectively. Then we unmerge the tokens after self-attention. The temporal merging and unmerging operations are the same as in \cite{li2024vidtome}.

\subsection{Text-Referenced Strategy}

\subsubsection{Dual Prompt Tuning Inversion}

PSLD \cite{rout2024solving} starts DDIM sampling from pure Gaussian noise latent and still requires 1000 steps of iteration for the reverse diffusion process. \cite{chung2022come,liu2023accelerating} have proved that starting from a better initialization can significantly reduce the number of sampling steps. We apply the Distortion Adaptive (DA) Inversion from \cite{liu2023accelerating} to convert $z_0=\mathcal{E}(A^Ty)$ into $z_{T'}$ to provide a good initialization.
\begin{equation}
\begin{split}
    z_{t+1} &= \sqrt{\alpha_{t+1}} \frac{z_t - \sqrt{1 - \alpha_t} \varepsilon_\theta}{\sqrt{\alpha_t}} \\
              &+ \sqrt{1 - \alpha_{t + 1} - \eta \beta_{t + 1}} \varepsilon_\theta + \sqrt{\eta \beta_{t + 1}}\epsilon_{t}'
\end{split}
\end{equation}
where $\eta$ ($0<\eta<1$) is a hyper-parameter, $\epsilon_{t}'$ is Gaussian noise and $\epsilon_{t}' \sim \mathcal{N}(0, 1)$, and $T'$ is set to 15. $z_{T'}$ is then employed for initialization. Although DDIM inversion is commonly used in image and video editing, it is not suitable for restoration. When inverting the latent from a degraded image into noisy latent, the predicted noise during the DDIM inversion process will deviate from the standard normal distribution, resulting in the deviation of the initialization from the predefined noise distribution, which generates unrealistic results.

Different from unconditional image diffusion models \cite{liu2023accelerating}, text-to-image latent diffusion models utilize the classifier-free guidance technique \cite{ho2022classifier} to amplify the effect of conditioned text.
\begin{equation}
\varepsilon_\theta(z_t,t,\mathcal{C},\varnothing) = w \cdot \varepsilon_\theta(z_t,t,\mathcal{C}) + (1-w) \cdot \varepsilon_\theta(z_t,t,\varnothing)
\end{equation}
For PSLD, $w=7.5$ is the default parameter. Inspired by \cite{mokady2023null}, we propose to optimize both conditional embeddings $\mathcal{C}$ and unconditional embeddings $\varnothing$ in the inversion to better encode the information of degraded images, namely dual prompt tuning inversion, which means automatically tuning the conditional text and unconditional null-text embeddings. The DA inversion with $w=0$ outputs a sequence of noisy latent codes $z_{T'}^{*}$, ..., $z_{0}^{*}$ where $z_{0}^{*}=z_0=\mathcal{E}(x_0^{degraded})$. We initialize $\Tilde{z}_{T'}=z_{T'}^*$ and perform the following optimization on the conditional embedding $\mathcal{C}$ with $w=7.5$ for the timestamps $t=T',...,1$.
\begin{equation}
\underset{\mathcal{C}_t}{\min}~||z^*_{t-1}-z_{t-1}(\Tilde{z}_t, t,\mathcal{C}_t, \varnothing)||_2^2 + \gamma_1||\mathcal{C}_t^i-\mathcal{C}_t^{i-1}||_2^2
\end{equation}
the second term constrains the $\mathcal{C}_t$ of frames $i$ and $i+1$ to have a similar value, which maintains temporal consistency. Through the above optimization, we can obtain the optimized conditional embeddings $\mathcal{C}$. Then we optimize for the unconditional embedding $\varnothing$, where the DA inversion with $w=1$ outputs a sequence of noisy latent codes $z_{T'}^{**}$, ..., $z_{0}^{**}$. We initialize $\overline{z}_{T'}=z_{T'}^{**}$ and perform the following optimization with $w=7.5$ for the timestamps
$t\!=\!T',...,1$:
\begin{equation}
\underset{\varnothing_t}{\min}~||z^*_{t-1}-z_{t-1}(\overline{z}_t, t,\mathcal{C}, \varnothing_t)||_2^2 + \gamma_2||\varnothing_t^i-\varnothing_t^{i-1}||_2^2
\end{equation}
where $\gamma_1$ and $\gamma_2$ are hyperparameters. Through the above optimization, we can get the optimized unconditional embeddings $\varnothing$. When applying inversion, we share the same noise $\epsilon_{t}'$ across all frames to further preserve temporal consistency. Through the proposed dual prompt tuning inversion, we can reduce sampling steps $T$ from 1000 to 250 with 60 additional inversion steps.

\subsubsection{Dual Prompt Tuning Sampling}

In the sampling process, we continue to optimize the conditional embeddings $\mathcal{C}$ and unconditional embeddings $\varnothing$ to improve the performance. Although the conditional text and unconditional null-text embeddings have been optimized in the inversion, the optimization focuses on encoding the input degraded frames to give a better initialization to accelerate the sampling. The optimized conditional and unconditional embeddings are not most suitable for the sampling process and need further optimization. Firstly, we initialize conditional and unconditional embeddings with the optimization results in the inversion. Then we optimize the conditional embeddings, unconditional embeddings, and the reverse diffusion result alternately for each timestamp $t = T, ..., 0$. In every timestep $t$ of sampling, a clean latent $\boldsymbol{\hat{z}}_0$ can be predicted from the noisy latent $\boldsymbol{z}_{t}$ by Eq. \ref{x0}, which is also influenced by $\mathcal{C}_t$ and $\varnothing_t$. For conditional embeddings $\mathcal{C}_t$, the optimization objective is:
\begin{equation}
\underset{\mathcal{C}_t}{\min}~||y - A(\mathcal{D}(\hat{z}_0(\mathcal{C}_t, \varnothing_t)))||_2^2 + \gamma_1'||\mathcal{C}_t^i-\mathcal{C}_t^{i-1}||_2^2
\end{equation}
the first term aligns $\mathcal{C}_t$ with the measurement $y$, the second term constrains the $\mathcal{C}_t$ of different frames to maintain temporal consistency. We fix $\mathcal{C}_t$ with the optimized result $\mathcal{C}_t^{*}$, and then optimize the unconditional embeddings $\varnothing_t$.
\begin{equation}
\underset{\varnothing_t}{\min}~||y - A(\mathcal{D}(\hat{z}_0(\mathcal{C}_t^{*}, \varnothing_t)))||_2^2 + \gamma_2'||\varnothing_t^i-\varnothing_t^{i-1}||_2^2
\end{equation}
the first term aligns $\varnothing_t$ with the measurement $y$, the second term constrains the $\varnothing_t$ of different frames to maintain temporal consistency. $\gamma_1'$ and $\gamma_2'$ are hyperparameters. With fixed optimized results $\mathcal{C}_t^{*}$ and $\varnothing_t^{*}$, the reverse diffusion results are optimized by the loss in PSLD. Due to the balance between computation cost and performance, we only optimize $\mathcal{C}$ and $\varnothing$ in $N$ iterations after $t < T_{dpts}$. $T_{dpts}$ is set to 5, and $N$ is set to 5.

The Dual Prompt Tuning Inversion and Sampling support both no-text (null-text) reference and text reference. For null-text reference, conditional embeddings are generated from null text.

\subsection{Image-Referenced Module}

First, we apply the above inversion to the reference image to obtain the noisy latent of the reference image. Then, in the sampling process, we transfer the texture from the reference image to the current frame using the proposed modules.

\subsubsection{Referenced Self-Attention}

Inspired by cross-frame attention in zero-shot video editing \cite{khachatryan2023text2video,yang2023rerender}, we propose to replace the original self-attention layers in the U-Net with referenced self-attention layers to transfer the texture of the referenced image to the current frame. For each spatial self-attention layer, the query, key and value $Q$, $K$, $V$ are obtained by linear projection of the feature $v_i$ of $I_i$. The corresponding spatial self-attention output is produced by $\textit{SelfAttn}(Q,K,V)=\textit{Softmax}(\frac{QK^T}{\sqrt{d}})\cdot V$.
\begin{equation}
  Q=W^Qv_i, K=W^Kv_i, V=W^Vv_i,
\end{equation}
where $W^Q$, $W^K$, $W^V$ are pre-trained matrices that project the inputs to query, key and value, respectively. We propose referenced self-attention which uses the key $K'$ and value $V'$ from both the current frame itself and the referenced image, allowing the current frame to pay attention to itself and the reference image simultaneously. The referenced self-attention output is produced by $\textit{Ref\_SelfAttn}(Q,K',V')=\textit{Softmax}(\frac{QK'^T}{\sqrt{d}})\cdot V'$.
\begin{equation}
  Q=W^Qv_i, K'=W^K[v_i, v_{ref}], V'=W^V[v_i, v_{ref}],
\end{equation}

\subsubsection{Referenced Token Merging}

To further transfer the texture of the referenced image to the current frame and reduce computational costs, we propose referenced token merging. We select the referenced image as the key image and merge the current frame into the referenced image based on token similarity between images. We also utilize the classifier in \cite{jeong2024accelerating} to classify pixels of the reference image as belonging to either smooth or texture-rich areas. Then we merge $\textbf{T}_{s}$ ($\textbf{T}_{t}$) from the referenced image and the current frame. After self-attention, we unmerge the merged tokens back into the original referenced image and current frame tokens. We alternately apply referenced self-attention and referenced token merging to the two self-attention modules in the transformer block of UNet, enhancing the fusion between the current frame and the referenced image for better texture transfer.

\section{Experiments}

\subsection{Datasets}

We evaluate our method on 2 video restoration tasks (video super-resolution, video deblur) and 1 video enhancement task (low-light video enhancement). For evaluation of video restoration, we collected 15 gt videos from commonly used test datasets for restoration (REDS4 \cite{nah2019ntire}, UDM10 \cite{yi2019progressive}, DAVIS \cite{pont20172017}). For evaluation of video enhancement, we collected 10 paired low-light and normal-light videos from the DID dataset \cite{fu2023dancing}, which was captured in the real world. Due to the slow sampling speed and the fact that a test video contains a lot of frames, we first center crop the frames along the shorter edge, and then resize them to 512$\times$512, which matches the image size of the text-to-image diffusion model. For degraded videos of restoration tasks, we follow the settings of the linear degradation operator from \cite{rout2024solving} and \cite{fei2023generative}. For video enhancement, we utilize the same degradation model as in \cite{fei2023generative} and optimize the parameters of the degradation model during the sampling process. Due to page limitations, we provide more details in the supplementary materials.

\begin{figure*}
    \centering
    \includegraphics[width=0.99\linewidth]{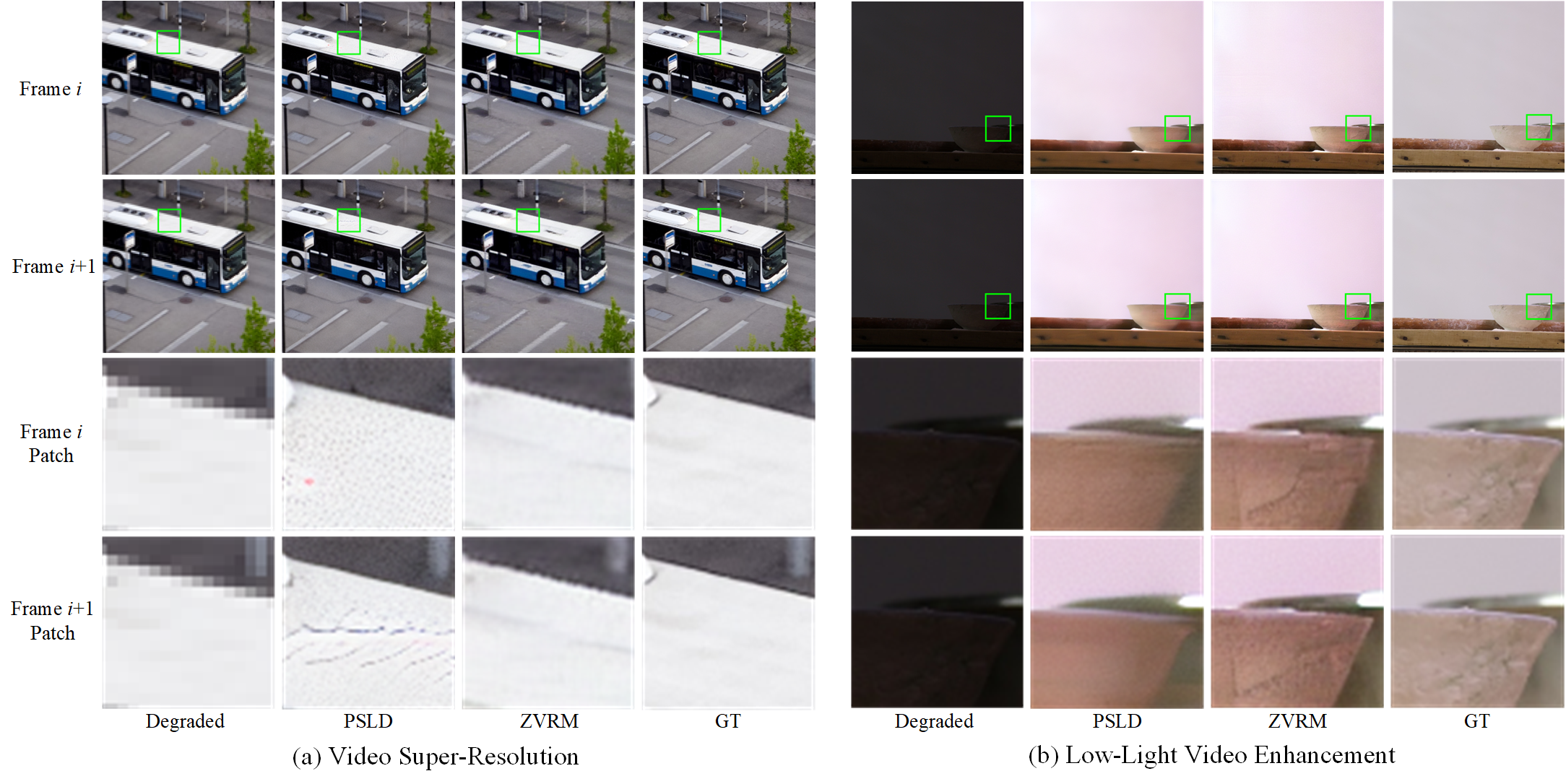}
    \caption{Visual quality comparison for video super-resolution and low-light video enhancement. Zoom in for better observation.}
    \label{fig:videosrenhance}
\end{figure*}



\begin{table}[t]
\centering
\caption{Quantitative comparison with state-of-the-art methods for 4$\times$ video super-resolution. The best results are highlighted in bold.}
\resizebox{0.60\textwidth}{10.5mm}{
\begin{tabular}{l|cccc}
\toprule
Methods                          & Backbone         & PSNR$\uparrow$  & SSIM$\uparrow$   & WE($10^{-2}$)$\downarrow$ \\
\hline
PSLD                             & SDv1.5          &24.85	          &0.6313	             &2.0854	               \\
ZVRM (no Ref.)                & SDv1.5          &25.16	          &0.6599	             &0.5003	               \\
ZVRM (text Ref.)              & SDv1.5          &25.32             &0.6642	             &0.4755	               \\
ZVRM (text Ref.+image Ref. 1) & SDv1.5          &25.61             &0.6785	             &\textbf{0.3871}	       \\
\hline
VISION-XL                        & SDXL          &25.91              &0.7487                 &1.5578                   \\
ZVRM (no Ref.)                & SDXL          &\textbf{26.13}	 &\textbf{0.7594}	     &0.9345	               \\
\bottomrule
\end{tabular}
}
\label{ComparisonVSR}
\end{table}

\begin{table}[t]
\centering
\caption{Quantitative comparison with state-of-the-art methods for low-light video enhancement. The best results are highlighted in bold.}
\resizebox{0.60\textwidth}{8.5mm}{
\begin{tabular}{l|cccc}
\toprule
Methods                           & Backbone     & PSNR$\uparrow$  & SSIM$\uparrow$   & WE($10^{-2}$)$\downarrow$ \\
\hline
PSLD                              & SDv1.5                   &18.17                      &0.8174	             &0.6895	               \\
ZVRM (no Ref.)                 & SDv1.5                   &18.40	                     &0.8303	             &0.2170	               \\
ZVRM (text Ref.)               & SDv1.5                   &18.45                      &0.8369	             &0.2032	                \\
ZVRM (text Ref.+image Ref. 1)  & SDv1.5                   &\textbf{18.79}	         &\textbf{0.8552}	     &\textbf{0.1924}	   \\
\bottomrule
\end{tabular}
}
\label{ComparisonVE}
\end{table}

\begin{table}[t]
\centering
\caption{Quantitative comparison of 4$\times$ video super-resolution using different reference image sources on the RealMCVSR dataset \cite{lee2022reference}. The best results are highlighted in bold.}
\resizebox{0.60\textwidth}{10.5mm}{
\begin{tabular}{l|cccc}
\toprule
Methods                           & Backbone       & PSNR$\uparrow$  & SSIM$\uparrow$   & WE($10^{-2}$)$\downarrow$ \\
\hline
PSLD                              & SDv1.5      &22.91	          &0.6241	             &1.9198	              \\
ZVRM (no Ref.)                 & SDv1.5      &23.14	          &0.6339	             &0.8215	              \\
ZVRM (image Ref. 1)            & SDv1.5      &\textbf{23.46}	  &\textbf{0.6552}	     &0.8069	              \\
ZVRM (image Ref. 2)            & SDv1.5      &23.33	          &0.6439	             &\textbf{0.7720}	      \\
ZVRM (image Ref. 3)            & SDv1.5      &23.37	          &0.6461	             &0.8013	              \\
\bottomrule
\end{tabular}
}
\label{ComparisonVSRDR}
\end{table}

\begin{table}[t]
\centering
\caption{Quantitative comparison with state-of-the-art methods for 4$\times$ blind video super-resolution on the DAVIS dataset. The best results are highlighted in bold.}
\resizebox{0.60\textwidth}{10.5mm}{
\begin{tabular}{l|cccc}
\toprule
Methods                                  & Backbone & PSNR$\uparrow$  & SSIM$\uparrow$   & WE($10^{-2}$)$\downarrow$  \\
\hline
DiffBIR                                  & SDv2.1 &24.47	          &0.6727	             &0.6907	              \\
DiffIR2VR                                & SDv2.1 &24.49	          &0.6718	             &0.6818	              \\
DiffBIR+ZVRD                             & SDv2.1 &24.55	          &0.6799	             &0.4923	              \\
DiffBIR+ZVRM (no Ref.)                & SDv2.1 &24.71	          &0.6839	             &0.4397	              \\
DiffBIR+ZVRM (text Ref.)              & SDv2.1 &24.80	          &0.6860	             &0.4375	              \\
DiffBIR+ZVRM (text Ref.+image Ref. 1) & SDv2.1 &\textbf{24.96}  &\textbf{0.6971}	     &\textbf{0.4288}	           \\
\bottomrule
\end{tabular}
}
\label{ComparisonBVSR}
\end{table}

\begin{figure}
    \centering
    \includegraphics[width=0.99\linewidth]{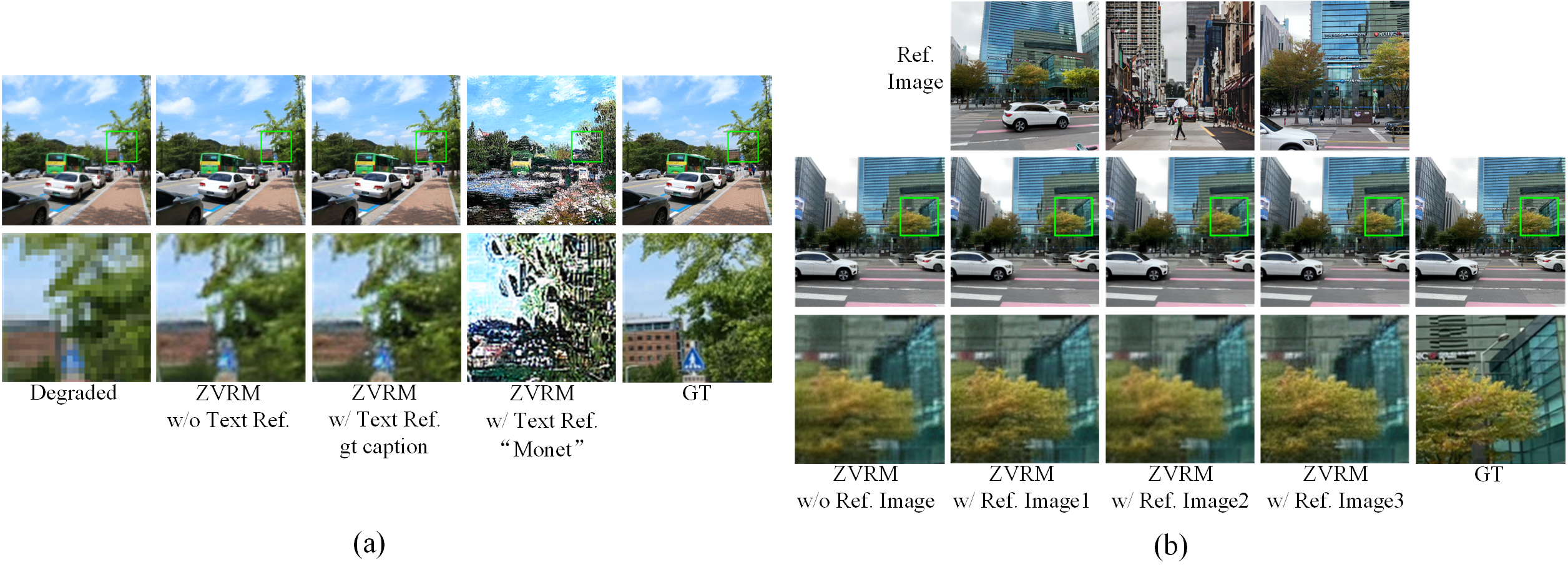}
    \caption{Different kinds of reference images for zero-shot video super-resolution on the RealMCVSR dataset.}
    \label{fig:videosrref}
\end{figure}

\begin{table}
\centering
\caption{Ablation study for Dual Prompt Tuning Inversion (DPTI), Dual Prompt Tuning Sampling (DPTS), Texture-Aware Spatial Token Merging (TSTM), Texture-Aware Temporal Token Merging (TTTM), Referenced Self-Attention (RSA), and Referenced Token Merging (RTM) on 4$\times$ video super-resolution task.}
\resizebox{0.60\textwidth}{14.5mm}{
\begin{tabular}{ccccccccc}
\toprule
\multicolumn{2}{l}{DPTT}            & $\times$  & $\checkmark$  & $\checkmark$    & $\checkmark$    & $\checkmark$  & $\checkmark$  & $\checkmark$   \\
\multicolumn{2}{l}{DPTS}             & $\times$  & $\times$      & $\checkmark$    & $\checkmark$    & $\checkmark$  & $\checkmark$  & $\checkmark$ \\
\multicolumn{2}{l}{TSTM}     & $\times$  & $\times$      & $\times$        & $\checkmark$    & $\checkmark$  & $\checkmark$  & $\checkmark$ \\
\multicolumn{2}{l}{TTTM}    & $\times$  & $\times$      & $\times$        & $\times$        & $\checkmark$  & $\checkmark$  & $\checkmark$ \\
\multicolumn{2}{l}{RSA}               & $\times$  & $\times$      & $\times$        & $\times$        & $\times$      & $\checkmark$  & $\checkmark$ \\
\multicolumn{2}{l}{RTM}                & $\times$  & $\times$      & $\times$        & $\times$        & $\times$      & $\times$      & $\checkmark$ \\ \hline
\multicolumn{2}{l}{PSNR$\uparrow$}              & 24.85  & 24.88  & 25.03  & 25.05   & 25.16    & 25.31    & 25.49      \\
\multicolumn{2}{l}{SSIM$\uparrow$}              & 0.6313 & 0.6501 & 0.6585 & 0.6572  & 0.6599   & 0.6654   & 0.6726      \\
\multicolumn{2}{l}{WE($10^{-2}$)$\downarrow$ }  & 2.0854 & 1.9646 & 1.0721 & 1.0710  & 0.5003   & 0.4622   & 0.4057      \\
\multicolumn{2}{l}{Steps$\downarrow$ }          & 1000   & 310    & 360    & 360     & 360      & 360      & 360      \\
\bottomrule
\end{tabular}
}
\label{Ablation}
\end{table}

\subsection{Comparison with State-of-the-art Methods}

We utilize three metrics to evaluate the restoration and enhancement quality. Besides the commonly used metrics PSNR and SSIM, we utilize Warping Error (WE) \cite{lai2018learning} to evaluate temporal consistency. Due to the page limitation, we provide more analyses of inference time in the supplementary materials. For zero-shot video restoration and enhancement, we mainly compare our method with PSLD and VISION-XL, PSLD is also our backbone. Since VISION-XL utilizes Stable Diffusion XL 1.0 (SDXL) instead of Stable Diffusion v1.5 (SDv1.5), like PSLD, we change our SD backbone to SDXL to ensure a fair comparison when comparing our method with VISION-XL. It is worth noting that our method is also a plug-and-play module, which can be applied to any diffusion-based model. And our method can also be applied to blind restoration tasks with complex real-world degradation. We transfer our method to the backbone DiffBIR for blind video super-resolution, comparing it with DiffIR2VR and ZVRD. 

For settings of no reference, we only give the null-text to our model. This is also the most common situation in the real world. Text references also exist in real world. For example, when people shoot video for sharing in the Internet, they sometimes also record a voiceover or introduction for the video. In addition, the videos on the Internet often have a description. For the restoration/enhancement task, users can also give a description for their desired output. To ease the evaluation, we follow \cite{kim2023regularization} and generate text from ground truth for the settings of text reference. To be specific, we feed the full resolution ground truth frame to GPT4o to generate reference text. We can also generate reference text from degraded frames or user-preferred style. However, this will cause the enhanced result deviates from the GT, which makes it difficult to evaluate the restoration methods. For image reference, we follow the settings of \cite{jiang2022reference} to ease the evaluation. To be specific, we select a GT frame that does not overlap with the test clip as the reference image. We name this kind of reference image Ref. 1. To explore the influence of different kinds of reference images, we also present results when using the clean image from retrieve (Ref. 2) and the image captured from another angle (Ref. 3) as reference image.

Table \ref{ComparisonVSR} and \ref{ComparisonVE} list the quantitative results on the evaluation data for video super-resolution and low-light video enhancement, respectively. It can be observed that our method ZVRM outperforms PSLD in all three metrics across the three tasks, even without any reference. For 4$\times$ video super-resolution, ZVRM (no Ref.) outperforms PSLD with a 0.31 dB gain in PSNR, has nearly only 1/4 of the WE value. For low-light video enhancement, ZVRM (no Ref.) outperforms PSLD with a 0.23 dB gain in PSNR, has nearly only 1/3 of the WE value. Text and image references (Ref. 1) can further enhance the performance. Our method also outperforms another zero-shot video restoration method, VISION-XL, in all three metrics. For 4$\times$ video super-resolution, ZVRM (no Ref.) outperforms VISION-XL with a 0.22 dB gain in PSNR, and has nearly half the WE value. This demonstrates the superiority of our method in terms of performance and temporal consistency. Due to page limitations, we provide the comparisons for video deblurring in the supplementary materials.


Fig. \ref{fig:videosrenhance} presents the visual comparison results on the evaluation data for video super-resolution and low-light video enhancement. For video super-resolution, the top area of the bus has different textures in the two frames of PSLD results. Our method can restore more temporally consistent results. For low-light video enhancement, it can be observed that PSLD results in a loss of many details, whereas our method can maintain more details and preserve temporal consistency. Due to page limitations, we provide more comparisons in the supplementary materials.

We further explore the influence of different kinds of reference texts and images. Fig. \ref{fig:videosrref} (a) shows the results when using a ground truth (gt) caption or a user-preferred style as reference text. When provided with a description of the user’s preferred style, such as “Monet,” as a reference text, our model can also generate the corresponding result with the desired style. Fig. \ref{fig:videosrref} (b) shows the results when using three kinds of reference images. The Ref. 2 image from Fig. \ref{fig:videosrref} (b) is retrieved from the DIV2K dataset by the retrieval system in \cite{guo2024refir}. Table \ref{ComparisonVSRDR} further provides the quantitative comparison results. It can be observed that every kind of reference image can improve the performance.

Following the settings of DiffIR2VR \cite{yeh2024diffir2vr}, we use DiffBIR as the backbone for blind video super-resolution and evaluate it on the DAVIS testing sets. We compare our method with DiffIR2VR, ZVRD, and the backbone DiffBIR. Table \ref{ComparisonBVSR} further provides the quantitative comparison results. It can be observed that our method achieves the best performance in all metrics. Due to page limitations, we provide visual quality comparison results in the supplementary materials.

\subsection{Ablation Study}

In this section, we perform an ablation study to demonstrate the effectiveness of the proposed dual prompt tuning inversion, dual prompt tuning sampling, referenced self-attention, referenced token merging, texture-aware spatial token merging, and temporal token merging. Taking video super-resolution as an example, Table \ref{Ablation} lists the quantitative comparison results in the evaluation data by adding these modules one by one. Our proposed modules can bring a 0.64 dB gain for PSNR, a 0.0403 gain for SSIM, nearly a 4/5 reduction for WE, and reduce the diffusion steps from 1000 to 360 (60 inversion steps and 300 sampling steps). By adding the dual prompt tuning inversion (null-text reference), the diffusion steps are accelerated from 1000 to 310 (60 inversion steps and 250 sampling steps) while slightly increasing performance. By replacing plain sampling with dual prompt tuning sampling (null-text reference), the performance of three metrics can be significantly improved with an additional cost of 50 sampling steps (0.15 dB for PSNR, nearly a 1/2 reduction for WE). Referenced self-attention and referenced token merging can bring gains of 0.15 dB and 0.18 dB for PSNR, respectively. Due to page limitations, we provide more ablation studies in the supplementary materials.

\section{Conclusion}

In this paper, we propose a novel framework for zero-shot video restoration using a pretrained text-to-image latent diffusion model and multi-modal references. Through the proposed dual prompt tuning inversion and sampling, the inference time can be reduced to nearly one-third of the original. The performance and temporal consistency can also be significantly strengthened. By using the proposed texture-aware video token merging, the temporal correlation between frames can be further utilized to improve the temporal consistency. We further propose referenced self-attention and referenced token merging to support image reference. Experimental results demonstrate the superiority of the proposed method in speed, performance, and temporal consistency.

{\small
\bibliographystyle{unsrt}
\bibliography{ms}
}

\end{document}


\maketitle
This supplementary file provides details which were not presented in the main paper due to page limitations. In the following, we first give the details for zero-shot video enhancement and inference time analyses. Then we present more comparison results and ablation study. And a demo for video results comparison is given. Finally, limitations and societal impact of our work are given.

\section{Learning degradation model for Zero-shot Video Enhancement}

Linear inverse problems in image restoration (IR) can be formulated as  
\begin{equation}
    y = Ax + n
\end{equation}
where $A$ is the linear degradation operator and $n$ is additive white Gaussian noise, the task is to restore the ground-truth image $x$ from the degraded image $y$. Following PSLD \cite{rout2024solving}, a latent constraint is applied in the reverse diffusion process to preserve the content between the generated result and the degraded image. The constraint loss is formulated as  
\begin{equation}
\begin{split}
\mathcal{L} &= \mathcal{L}_{rec} + \gamma_1\mathcal{L}_{reg} \\
\mathcal{L}_{rec} &= \|y - A(\mathcal{D}(\hat{z}_0))\|_2^2 \\
\mathcal{L}_{reg} &= \|\hat z_0 - \mathcal{E}({A^Ty + (I - A^TA) \mathcal{D}(\hat z_0))}\|_2^2
\end{split}
\label{loss}
\end{equation}
where $\mathcal{L}_{rec}$ directly constrains the content, $\mathcal{L}_{reg}$ penalizes latents that are not fixed-points of the composition of the decoder-function with the encoder-function, ensuring that the generated sample remains on the manifold of real data. For zero-shot video enhancement, we utilize the degradation model in \cite{fei2023generative} and optimize the parameters of the degradation model in the sampling process. The degradation model can be formulated as follows:
\begin{equation}
    \boldsymbol{y}=f \boldsymbol{\mathcal{D}(\hat{z}_0)}+\boldsymbol{\mathcal{M}}, 
\label{eq:light}
\end{equation}
where the factor $f$ is a scalar and the mask $\boldsymbol{\mathcal{M}}$ is a matrix of the same dimension as $\boldsymbol{\mathcal{D}(\hat{z}_0)}$. We optimize $f$ and $\boldsymbol{\mathcal{M}}$ using gradient descent during the sampling process.

\section{Inference Time Analyses}

The algorithm in \ref{alg} shows our entire reverse diffusion process. All experiments were conducted on a 48G A40 GPU. Taking the backbone PSLD as an example, each sampling step takes approximately 1.424 seconds without any reference. Our dual prompt tuning inversion and sampling can reduce the total sampling steps from 1000 to 360, reducing the total inference time from 23min44s to 8min32s. Our texture-aware video token merging can further reduce the cost of each sampling step from 1.424s to 1.184s, reducing the total inference time from 8min32s to 7min6s.

\begin{algorithm}[!t]
\renewcommand{\arraystretch}{0.75}
\caption{Sampling process of the proposed ZVRM (without reference).}
\begin{algorithmic}[1] 
\REQUIRE Corrupted video $\{I_i\}_{i=0}^{N-1}$, conditional embeddings $\mathcal{C}$, unconditional embeddings $\varnothing$, latent $z$, autoencoder $\mathcal{E}$ and $\mathcal{D}$
\ENSURE Restored video $\{I'_i\}_{i=0}^{N-1}$ 
\STATE $z_0^{inversion}=\mathcal{E}(I)$
\FOR{$t=0$ to $T'-1$} 
\FOR{$i=0$ to $N-1$}
\STATE      OptimizeInversion($\Tilde{z}_t$, $t$, [$\mathcal{C}_t^{i-1}$, $\mathcal{C}_t^{i}$], $\varnothing$)
\ENDFOR 
\ENDFOR 
\FOR{$t=0$ to $T'-1$} 
\FOR{$i=0$ to $N-1$}
\STATE      OptimizeInversion($\overline{z}_t$, $t$, $\mathcal{C}$, [$\varnothing_t^{i-1}$, $\varnothing_t^{i}$])
\ENDFOR
\ENDFOR 
\FOR{$t=T-1$ to 0}
\FOR{$i=0$ to $N-1$}
\FOR{$k=1$ to K} 
\STATE      OptimizeSampling($z_t$, $y$, [$\mathcal{C}_t^{i-1}$, $\mathcal{C}_t^{i}$])
\ENDFOR
\FOR{$k=1$ to K} 
\STATE      OptimizeSampling($z_t$, $y$, $\mathcal{C}_t^{*}$, [$\varnothing_t^{i-1}$, $\varnothing_t^{i}$])
\ENDFOR
\STATE OptimizeSampling($z_t$, $y$, $\mathcal{C}_t^{*}$, $\varnothing_t^{*}$)
\STATE $z_t^{i}$ = TextureAwareVideoTokenMerging($z_t^{i-1}$, $z_t^{i}$)
\ENDFOR
\ENDFOR 
\STATE $I' = \mathcal{D}(z_0^{sampling})$

\end{algorithmic} 
\label{alg}
\end{algorithm}

\section{Comparison with State-of-the-art Methods}

Table \ref{ComparisonVDB} lists the quantitative results on the evaluation data for zero-shot video deblurring. It can be observed that our method (ZVRM, no Ref.) outperforms PSLD by 0.74 dB in terms of PSNR, and has nearly only 1/8 of the WE value. Text and image references can further improve the performance.

\begin{figure*}
    \centering
    \includegraphics[width=0.99\linewidth]{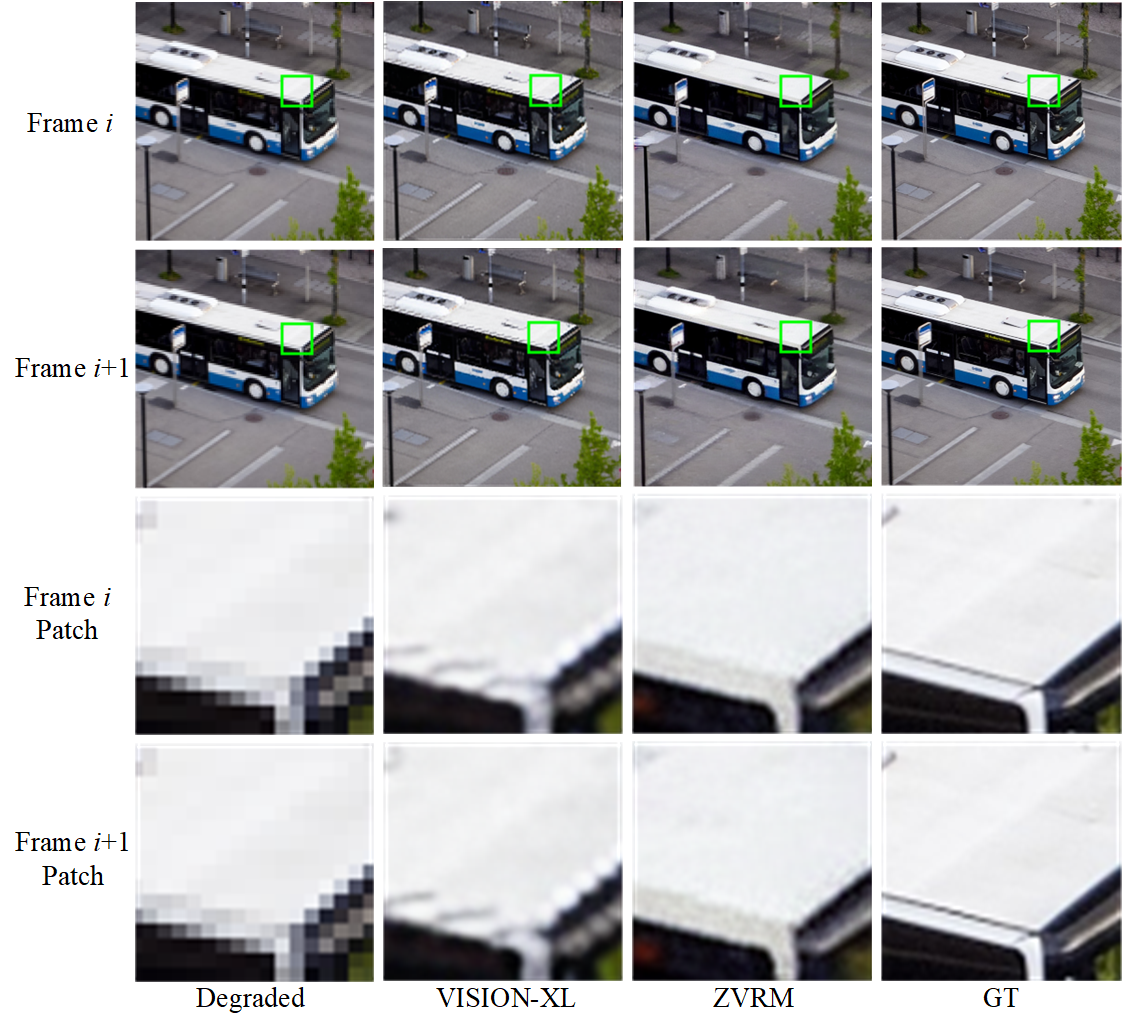}
    \caption{Visual quality comparison for video super-resolution using an SDXL backbone. Zoom in for better observation.}
    \label{fig:videosr}
\end{figure*}

\begin{figure*}
    \centering
    \includegraphics[width=0.99\linewidth]{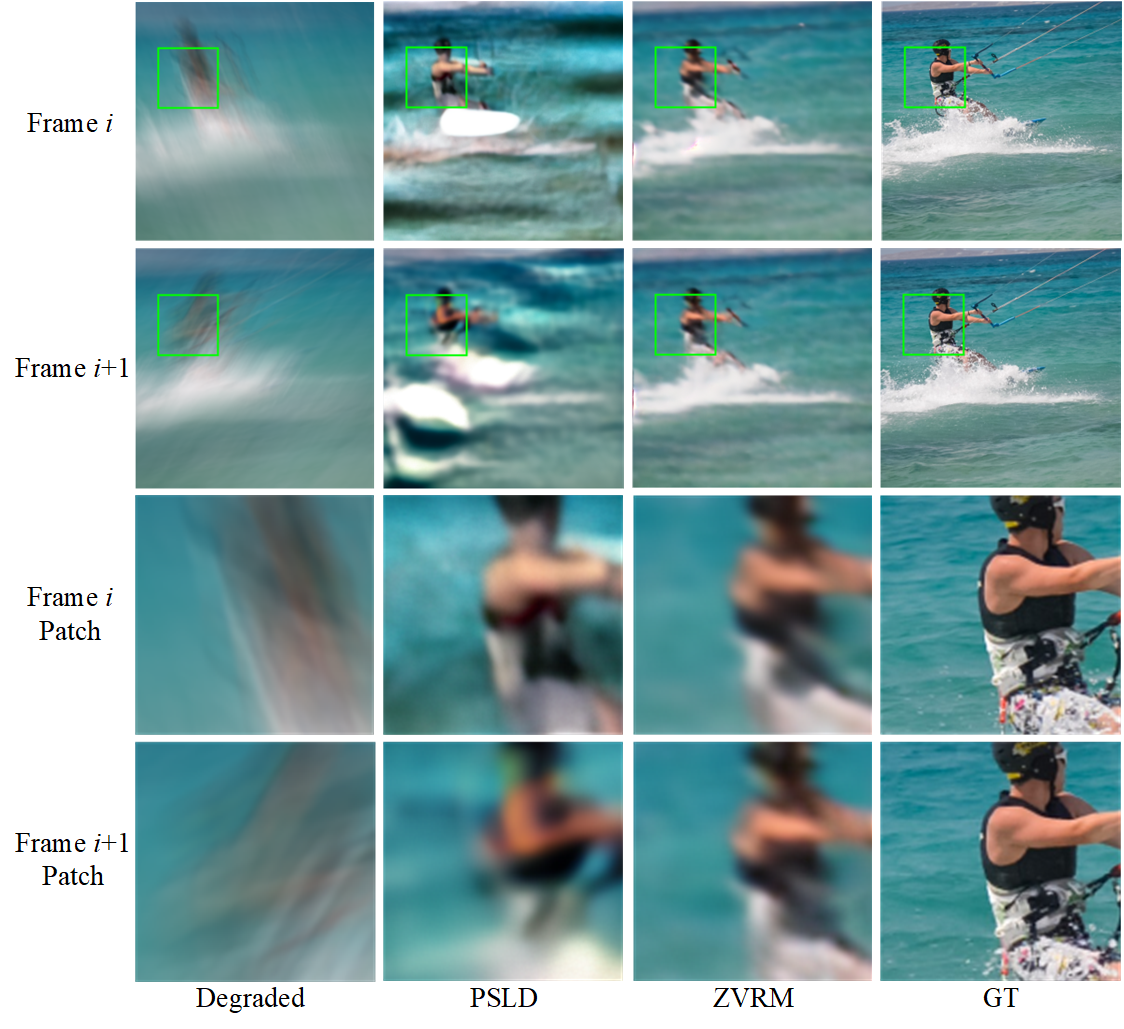}
    \caption{Visual quality comparison for video deblurring. Zoom in for better observation.}
    \label{fig:videodeblur}
\end{figure*}

Fig. \ref{fig:videosr} and \ref{fig:videodeblur} present the visual comparison results on the evaluation data for zero-shot video super-resolution and deblurring. For video super-resolution, we change the stable diffusion backbone from SDv1.5 to SDXL, and compare our method with VISION-XL. It can be observed that there are unpleasant artifacts in the edges of VISION-XL results. Our method (ZVRM, no Ref.) can restore clearer edges. For video deblurring, it can be observed that the people in the PSLD results have different body parts on the two frames. Our method (ZVRM, no Ref.) has better temporal consistency for the people's bodies and waves.

\begin{table}[t]
\centering
\caption{Quantitative comparison with state-of-the-art methods for video deblurring. The best results are highlighted in bold.}
\resizebox{0.70\textwidth}{11.0mm}{
\begin{tabular}{l|cccc}
\toprule
Methods                              & Backbone   & PSNR$\uparrow$  & SSIM$\uparrow$   & WE($10^{-2}$)$\downarrow$ \\
\hline
PSLD                                 & SDv1.5        &19.33                     &0.4755	             &5.4533	               \\
ZVRM (no Ref.)                    & SDv1.5        &20.07                     &0.4852	             &0.8183                   \\
ZVRM (text Ref.)                  & SDv1.5        &20.21	                    &0.4831	             &0.7697	               \\
ZVRM (text Ref.+image Ref. 1)     & SDv1.5        &\textbf{20.45}	        &\textbf{0.4910}	 &\textbf{0.7009}          \\
\bottomrule
\end{tabular}
}
\label{ComparisonVDB}
\end{table}

\begin{figure}
    \centering
    \includegraphics[width=0.99\linewidth]{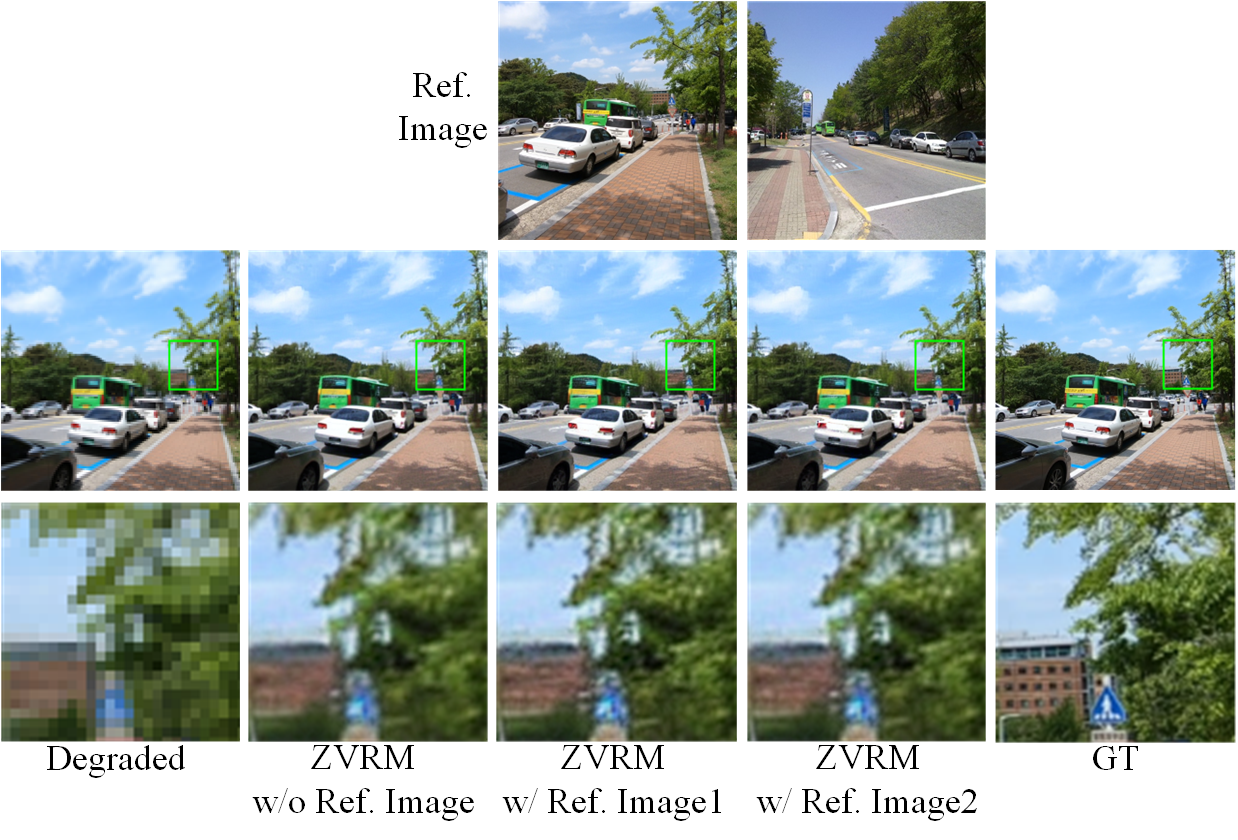}
    \caption{Different kinds of reference images for zero-shot video super-resolution.}
    \label{fig:videosrimageref1}
\end{figure}

Fig. \ref{fig:videosrimageref1} shows more results when using different kinds of reference images. The Ref. 2 image in Fig. \ref{fig:videosrimageref1} is retrieved by Google's image retrieval system.

\begin{figure}
    \centering
    \includegraphics[width=0.99\linewidth]{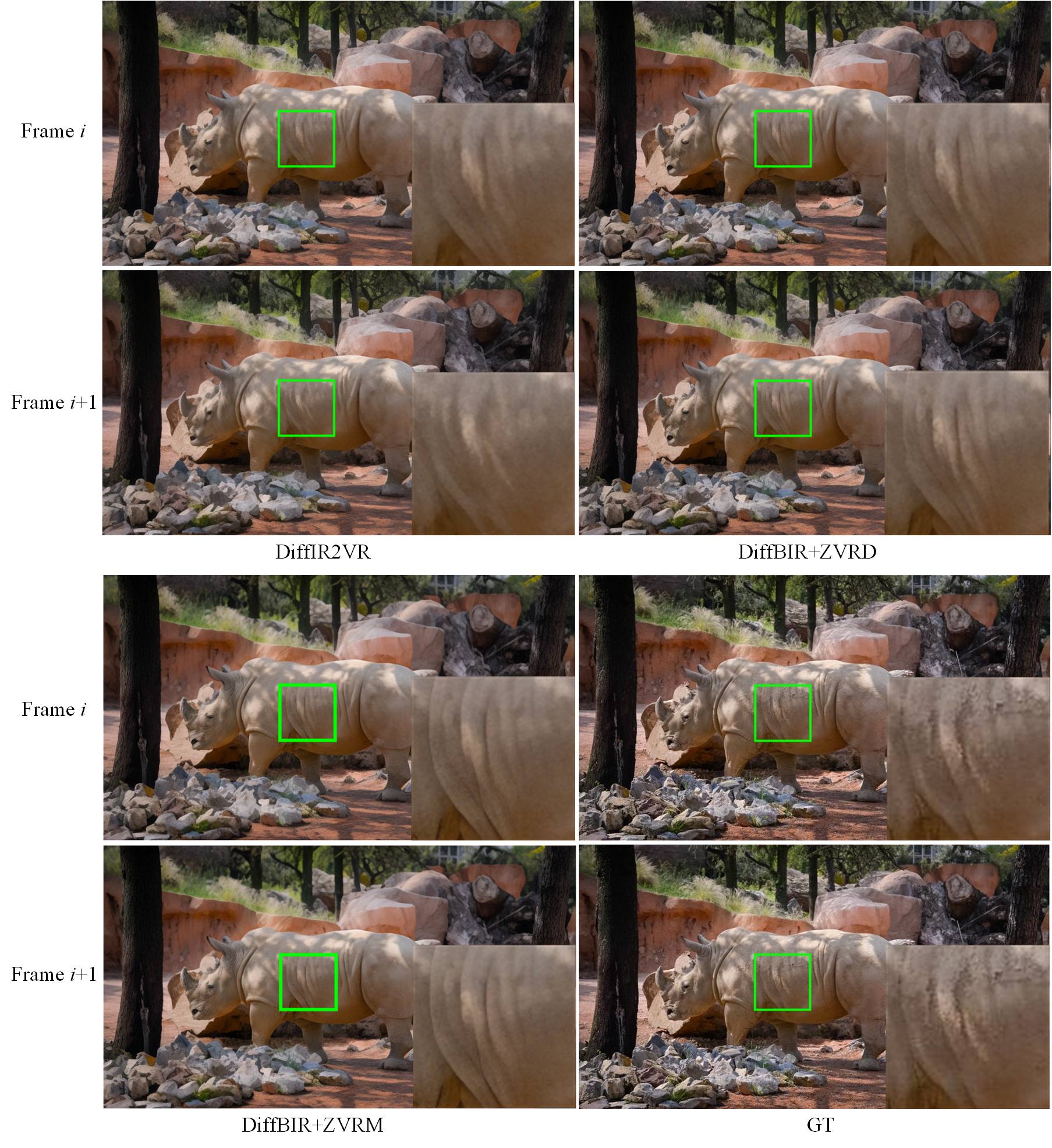}
    \caption{Visual quality comparison for blind video super-resolution. Zoom in for better observation..}
    \label{fig:videobsr}
\end{figure}

Our method can also be applied to blind restoration tasks with complex real-world degradation. Fig. \ref{fig:videobsr} gives the visual quality comparison results, and our method can restore sharp and temporal-consistent details. We also present a video demo to present the temporal consistency of our method (ZVRM, no Ref.).

\section{Ablation Study}

We provide more ablation studies on the modules of Dual Prompt Tuning Inversion (DPTI) and Dual Prompt Tuning Sampling (DPTS). Table \ref{AblationIS} lists the quantitative comparison results in evaluation data by adding the modules one by one. Optimizing conditional and unconditional embeddings both brings gains.

\begin{table}
\centering
\caption{Ablation study for dual prompt tuning inversion, dual prompt tuning sampling, spatial-temporal self-attention and temporal latent fusion on 4$\times$ video super-resolution task. Cond. and Uncond. respectively represent the optimization of conditional and unconditional embedding.}
\resizebox{0.60\textwidth}{16.5mm}{
\begin{tabular}{ccccccc}
\toprule
\multirow{2}{*}{DPTI} &Cond.              & $\times$     & $\checkmark$    & $\checkmark$ & $\checkmark$  & $\checkmark$       \\\cline{2-7}
                                              &Uncond.            & $\times$     & $\times$        & $\checkmark$ & $\checkmark$  & $\checkmark$      \\ \hline         
\multirow{2}{*}{DPTS}  &Cond.             & $\times$     & $\times$        & $\times$     & $\checkmark$  & $\checkmark$         \\\cline{2-7}
                                              &Uncond.            & $\times$     & $\times$        & $\times$     & $\times$      & $\checkmark$        \\\hline
\multicolumn{2}{l}{PSNR$\uparrow$}              & 24.85   &24.87   & 24.88  &24.96     & 25.03  \\
\multicolumn{2}{l}{SSIM$\uparrow$}              & 0.6313  &0.6457  & 0.6501 &0.6552    & 0.6585 \\
\multicolumn{2}{l}{WE($10^{-2}$)$\downarrow$ }  & 2.0854  &2.0197  & 1.9646 &1.3880    & 1.0721 \\
\multicolumn{2}{l}{Steps$\downarrow$ }          & 1000    &310     & 310    &335       & 360    \\
\bottomrule
\end{tabular}
}
\label{AblationIS}
\end{table}

\section{Limitations}

We would like to point out that our method also has some limitations. Taking the backbone PSLD as an example, although our method has reduced the total sampling steps from 1000 to 360, it still takes 7 minutes and 6 seconds to generate an image on an A40 GPU. We will further accelerate the sampling in future work.

\section{Societal Impact}

Video restoration and enhancement are widely used in video surveillance, video photography, etc. However, training a latent diffusion model for video restoration and enhancement requires substantial resources (several high-memory and fast GPUs). Therefore, in this paper, we propose a training-free method that utilizes a pre-trained text-to-image latent diffusion model. Our work can reduce the cost of experiments, accelerate research on video restoration and enhancement, and better serve relevant areas. At the same time, this kind of training-free video restoration and enhancement may bring more challenges in defending against attacks. We will further improve our method for deployment in critical security-sensitive systems.

{\small
\bibliographystyle{unsrt}
\bibliography{supplement}
}